\documentclass[letterpaper, 10 pt, conference]{ieeeconf}
\IEEEoverridecommandlockouts    %
\usepackage{graphics}           
\usepackage{times}              
\usepackage{amsmath}            
\usepackage{amssymb}            
\usepackage{graphicx}
\usepackage[]{algorithm2e}
\usepackage{booktabs}
\usepackage{color}
\usepackage{listings}
\usepackage{subfiles}
\usepackage{hyperref}
\usepackage[nocompress]{cite} %
\definecolor{instructioncolor}{rgb}{.5,.5,.5}

\usepackage[font=small]{caption}

\def\secref#1{Sec.~\ref{#1}}
\def\figref#1{Fig.~\ref{#1}}
\def\tabref#1{Tab.~\ref{#1}}
\def\eqref#1{Eq.~(\ref{#1})}

\makeatletter
\usepackage{xspace}
\DeclareRobustCommand\onedot{\futurelet\@let@token\@onedot}
\def\@onedot{\ifx\@let@token.\else.\null\fi\xspace}
\def\eg{e.g\onedot} 
\def\ie{i.e\onedot}

\def\etal{{et al}\onedot}
\makeatother

\def\etalcite#1{\etal~\cite{#1}}

\usepackage{array}
\newcolumntype{L}[1]{>{\raggedright\let\newline\\\arraybackslash\hspace{0pt}}m{#1}}
\newcolumntype{C}[1]{>{\centering\let\newline\\\arraybackslash\hspace{0pt}}m{#1}}
\newcolumntype{R}[1]{>{\raggedleft\let\newline\\\arraybackslash\hspace{0pt}}m{#1}}

\newcommand{\RR}{\mathbb{R}}

\newcommand{\norm}[1]{\lVert#1\lVert}

\renewcommand{\b}[1]{\mbox{\boldmath$#1$}}

\renewcommand{\v}[1]{{\b #1}} 

\newcommand{\m}[1]{{\mbox{{\sffamily\slshape{#1\/}}}}}

\usepackage{tabularx}
\usepackage{subcaption}
\usepackage{mathtools}
\usepackage{multirow}
\usepackage{flushend}

\newcolumntype{C}{>{\centering\arraybackslash}X}

\title{\LARGE \bf Efficient LiDAR Bundle Adjustment for Multi-Scan Alignment \\Utilizing Continuous-Time Trajectories }

\author{Louis Wiesmann \and Elias Marks \and Saurabh Gupta \and Tiziano Guadagnino \and Jens Behley \and Cyrill Stachniss%
  \thanks{All authors are with the Center for Robotics, University of Bonn, Germany. Cyrill Stachniss is additionally with the Department of Engineering Science at the University of Oxford, UK, and with the Lamarr Institute for Machine Learning and Artificial Intelligence, Germany.}%
  \thanks{This work has partially been funded 
  by the Deutsche Forschungsgemeinschaft (DFG, German Research Foundation) under Germany's Excellence Strategy, EXC-2070 -- 390732324 -- PhenoRob,
  by the European Union’s Horizon Europe research and innovation programme under grant agreement No~101070405~(DigiForest), 
  and
  by the German Federal Ministry of Education and Research (BMBF) in the project ``Robotics Institute Germany'', grant No.~16ME0999.
  }%
}

\begin{document}
\thispagestyle{empty}
\pagestyle{empty}
\maketitle

\begin{abstract}
  Constructing precise global maps is a key task in robotics and is required for localization, surveying, monitoring, or constructing digital twins.
  To build accurate maps, data from mobile 3D LiDAR sensors is often used.
  Mapping requires correctly aligning the individual point clouds to each other to obtain a globally consistent map.
  In this paper, we investigate the problem of multi-scan alignment to obtain globally consistent point cloud maps.
  We propose a 3D LiDAR bundle adjustment approach to solve the global alignment problem and jointly optimize the available data.
  Utilizing a continuous-time trajectory allows us to consider the ego-motion of the LiDAR scanner while recording a single scan directly in the least squares adjustment.
  Furthermore, pruning the search space of correspondences and utilizing out-of-core circular buffer enables our approach to align thousands of point clouds efficiently.
  We successfully align point clouds recorded with a handheld LiDAR, as well as ones mounted on a vehicle, and
  are able to perform multi-session alignment.
\end{abstract}

\section{Introduction}
\label{sec:intro}

Precise and consistent maps are crucial for localization, surveying, monitoring, or to generate digital twins.
3D LiDAR sensors allow for obtaining point clouds of the sensor's surroundings with millions of points per second.
Moving the sensor through the scene allows for capturing large areas in relatively short time.
To obtain a globally consistent point cloud map from these measurements, one needs to register and align all scans and transform each point into a unified map coordinate frame.

In this paper, we investigate the problem of registering large sets of point clouds consistently, namely tackling LiDAR bundle adjustment.
Matching two point clouds with each other is a well-studied problem in literature, where the most prominent method is the iterative closest point~(ICP) algorithm~\cite{besl1992pami} and its variants~\cite{rusinkiewicz2001dim}.
Applying ICP to successive scans, often done in an online fashion, is called LiDAR odometry~\cite{shan2018iros,vizzo2023ral,wang2021iros-fflo}.
This usually provides relatively well-aligned point clouds for short to medium-sized sequences.
However, odometry systems suffer from drift, often leading to bad globally aligned maps.
To overcome the problem, one can incorporate loop closures, which leads to an online SLAM system, or exploit global positioning information, \eg, GNSS data, in a pose-graph.
The optimization skews the whole trajectory to reduce the overall alignment error.
This usually leads to well-aligned maps, but may lack local consistency,
as the pose graph redistributes the errors along the whole map proportional to the uncertainty of the poses.
One can try to increase the uncertainty for the erroneous poses, but this requires knowledge about which pose or through which observations the error was introduced, however this information is usually not available.

LiDAR bundle adjustment and offline SLAM estimate the trajectory jointly with the map, and often uses the aforementioned systems to obtain an initial guess.
The big advantage over online SLAM and odometry systems is that LiDAR bundle adjustment does not have to work incrementally, but can utilize all the available information at once.
LiDAR bundle adjustment does not operate in real-time due to the large compute needed to find data associations and for optimizing the map and poses globally.
However, for several applications, the quality of the resulting map is key and more important than the runtime.

\begin{figure}[t]
  \centering
  \includegraphics[width=0.95\linewidth]{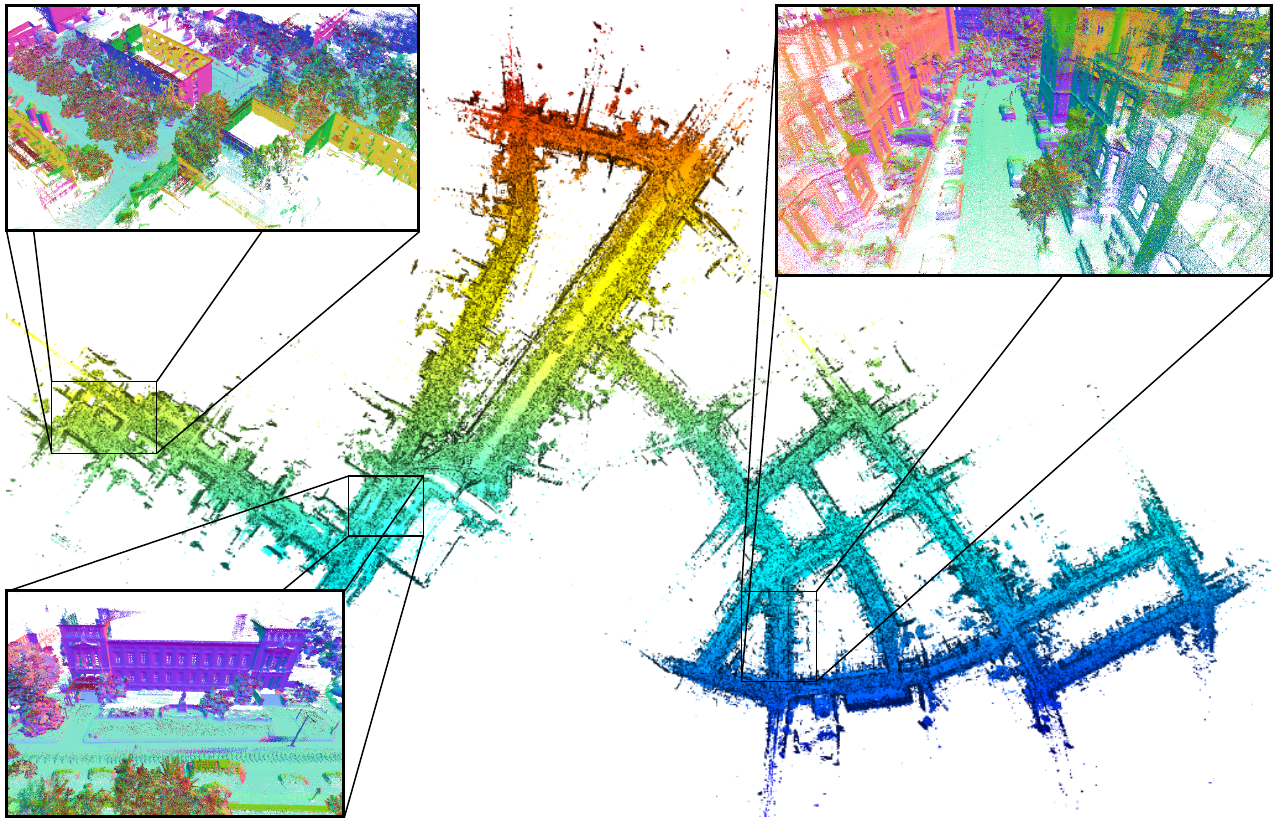}
  \caption{We propose a LiDAR bundle adjustment approach for aligning multiple point clouds. As shown here, our approach is able to generate a globally and locally aligned point cloud map. The color of the full point cloud is based on the latitude, while the points of the close-ups are colorized based on normals.  }
  \label{fig:motivation}
\end{figure}

The main contribution of this paper is a LiDAR bundle adjustment approach to globally align a large set of point clouds.
Starting from a set of scans and an initial pose estimate, our approach tries to align all the point clouds with each other.
We jointly optimize thousands of point clouds, resulting in a single big least squares adjustment at the end.
One exemplary result is shown in \figref{fig:motivation}, where the point cloud map after our alignment is depicted.
Utilizing a continuous-time trajectory allows us to model the motion distortion of the scans internally without the need for odometry or IMU.
Thus, each individual beam in a scan will be treated according to its measurement time.
The usage of an out-of-core circular buffer and pruning the search space of correspondences allows us to run our approach on thousands of point clouds.
Our approach is able to perform single-session as well as multi-session alignment.

In sum, we make three key claims:
Our approach is able to
(i) estimate a continuous-time trajectory of a 3D LiDAR scanner using LiDAR bundle adjustment,
(ii) which allows for the efficient construction of high-quality large-scale maps, and
(iii)  is able to handle multi-session alignment to obtain a unified global map.
These claims are backed up by the paper and our experimental evaluation.

\section{Related Work}
\label{sec:related}
Point cloud alignment is a well-studied problem in literature.
Most approaches that tackle the problem are based to some extent on the iterative closest point (ICP) algorithm \cite{besl1992pami}.
ICP estimates the transformation between two point clouds iteratively by a set of point correspondences, where in each iteration the correspondences are updated by searching for each point for the closest point in the other point cloud.
The classical point-to-point ICP~\cite{besl1992pami} requires sometimes many iterations to converge, which is why different error metrics, starting from point-to-plane~\cite{chen1991iros} up to entity-to-entity~\cite{segal2009rss} relations, have been proposed to speed up the optimization.
The closest point assumption relies on a good initial pose estimate to find reliable correspondences, which are necessary to converge to the correct solution.
Feature-based correspondence search~\cite{gelfand2005robust,johnson1999ivc,kerl2013icra,park2017iccv} can provide reliable data associations and can make them more robust against the quality of the initial guess.
However, the problem with feature-based matching is finding a suitable feature extractor that works reliably in different environments and under different conditions.
Therefore, our approach uses the more classical closest point assumption for all 3D points to be more independent of the sensor data.

Searching for correspondences is often the most time-consuming part.
Acceleration structures, like hash maps~\cite{niessner2013siggraph,pan2024tro,vizzo2023ral}, voxel grids~\cite{whelan2014ijrr}, octrees~\cite{steinbruecker2014icra}, or KDTrees~\cite{zhang2014rss}, as well as projective correspondence search~\cite{behley2018rss,newcombe2011ismar,whelan2015rss,keller2013threedv} are commonly used to speed up the search time.

Aligning sequential LiDAR data from classical multi-beam LiDAR is tackled by LiDAR odometry and graph-based simultaneous localization and mapping (SLAM).
The procedures usually utilize ICP to match successive scans.
Many different types of representations have been investigated: point-clouds~\cite{vizzo2023ral,zhang2014rss}, voxels~\cite{newcombe2011ismar}, normal distance transforms~\cite{biber2003iros, einhorn2015ras,stoyanov2012ijrr,das2013icra-3sru,saarinen2013iros-f3mi}, surfels~\cite{behley2018rss,droeschel2018icra,wang2021icra,stueckler2014vcir,park2018icra-elfd}, and implicit representations~\cite{deschaud2018icra,pan2024tro}.
While LiDAR odometry tries to solve the local alignment of successive scans, SLAM also tries to obtain globally consistent maps by incorporating loop closures~\cite{kaess2008tro}.
Since the LiDAR measures while in motion, the scans are often distorted, especially when moving at higher speeds. To deskew the scans, one can use a motion model~\cite{vizzo2023ral,zhang2014rss}, measure the sensor's motion using IMUs~\cite{shan2020iros,wei2022tro}, or estimate a continuous-time trajectory~\cite{dellenbach2022icra,droeschel2018icra,park2018icra-elfd,park2022tro}.

Classical photogrammetric bundle adjustment~\cite{triggs1999iccv,schneider2012isprs} utilizes images to jointly estimate poses as well as the 3D points.
Similarly, LiDAR bundle adjustment\cite{benjemaa1998eccv,bergevin1996tpami,liu2021ral}, and offline SLAM~\cite{dai2019rs,thrun2006ijrr} aim at globally aligning a set of point clouds into a consistent map.
Our approach belongs to this class of methods and tries to optimize the whole trajectory and the map together.
Due to the availability and often necessity of processing a large number of point clouds, approaches often try to reduce the complexity of the problem in a divide-and-conquer fashion.
Some approaches utilize feature-based point-line/plane ICP~\cite{liu2021ral} while others try to optimize directly to obtain a smooth and thin surface~\cite{li2024arxiv,skuddis2024arxiv}.
Di Giammarino~\etalcite{giammarino2023ral} aim to minimize a photometric error by utilizing the intensity channel of the LiDAR.
Projective data association techniques find correspondences, while they use image pyramids to make the approach more robust against the initial alignment error.
Liu~\etalcite{liu2023ral} tackle the problem in a divide-and-conquer manner and hierarchically optimizes the trajectory.
The main problem with hierarchical splitting is to tune the trade-off between local consistency and global consistency accordingly.

Our method does not make any assumptions about the scan pattern of the laser, nor relies on feature-based matching.
We utilize a continuous-time trajectory to have a pose for every beam recorded.
To make our approach computationally feasible, we instead sample corresponding scans, utilize out-of-core processing, as well as the parallel processing capabilities of modern GPUs.

\section{LiDAR Bundle Adjustment}
\label{sec:main}

\subsection{Problem Definition}
Modern 3D LiDAR sensors, e.g., rotating multi-beam LiDAR, provide point clouds of the sensor's surrounding.
Assuming a dataset $\mathcal{D} = \{\mathcal{P}_i\}$ of $N_{\text{scans}}$ point clouds, where each point cloud $\mathcal{P}_i= \{({\v p_j}, t_j)\}$ consists of $N$ points $\v p_j \in \RR^3$ and their corresponding timestamps~$t_j\in \RR$ at which the point is measured.
Each point $\v p_j$ is located in the local coordinate frame of the LiDAR at timestamp~$t_j$.
To globally align the point clouds~$\mathcal{P}_i$, we need to find the transformations~$\mathcal{T} = \{(\m R_{t_j},\v t_{t_j}) \mid \forall j\}$ given by the rotations~\mbox{$\m R_{t_j} \in SO(3)$} and translations~$\v t_{t_j} \in \RR^3$ for each beam that transforms the point $\v p_j$
from the local coordinate frame at timestamp~$t_j$ into a global coordinate frame:
\begin{equation}
  \label{eq:transform}
  \hat{\v p}_j= \m R_{t_j} \v p_j +\v t_{t_j},
\end{equation}
where $\hat{\v p}_j$ denotes a point in the global coordinate frame.
Point clouds from common LiDAR sensors, \eg, Velodyne HDL-64E, Ouster1-128 or similar, contain points measured at different points in time due to their measuring process.
This requires either deskewing the scans (\eg, using a constant velocity model \cite{vizzo2023ral}, or an estimate of the sensor's motion provided by an IMU~\cite{wu2024icra}) or to estimate a continuous trajectory~\cite{dellenbach2022icra,droeschel2018icra} over time to actually model the continuous motion of the sensor in space.
We utilize a continuous-time trajectory inspired by Dellenbach~\etalcite{dellenbach2022icra} and interpolate the poses between the beginning and end of each scan using spherical linear interpolation (slerp) for the rotation and linear interpolation for the translation.
Applying the formulas for the continuous-time trajectory results the transformations 
\begin{align}
  \m R_{t_i} & = \text{slerp}(\m R_{t_{b(j)}},\m R_{t_{e(j)}}, \alpha_i), \\
  \v t_{t_i} & = (1-\alpha) \v t_{t_{b(j)}}    + \alpha \v t_{t_{e(j)}},
\end{align}
where $\alpha_j$ is the fraction of time between the start $t_{b(j)}$ and end $t_{e(j)}$ of the corresponding scan for a given point $\v p_j$:
\begin{align}
  \alpha_j & = \frac{t_j - t_{b(j)}}{t_{e(j)}-t_{b(j)}}.
\end{align}

We assume the end pose of one scan to be the start pose of the successive scan.
Our objective function that we try to minimize is the squared distance of all corresponding points between each scan $\mathcal D$
\begin{equation}
  E = \sum_{j} \sum_{\v c \in C(\v p_j)} d(\hat{\v p}_j, \hat{\v p}_c)^2,
\end{equation}
where $d(\cdot)$ is a distance function (like point-to-point or point-to-plane), and ${C(\v p_j) = \{\v p_c \mid \hat{\v p}_c \equiv \hat{\v p}_j \}}$ is the set of points that correspond to the same location as the point $\v p_j$.
For the point-to-plane distance function, we yield an error function~$e_j$ for an arbitrary point $\v p_j$ as
\begin{equation}
  \label{eq:error_function}
  e_j = \hat{\v n}_c^\top (\hat{\v p}_j - \hat{\v p}_c),
\end{equation}
where $\hat{\v n}_c=\m R_{t_c} \v n_c$ is the normal of the point $\v p_c$ computed based on the local neighborhood within its scan.
The derivatives \cite{dellenbach2022icra} for the transformation of point $\hat{\v p}_j$ are given by
\begin{align}
  \v J_{\m R_{t_{b(j)}}} & = -(1-\alpha_j)  \hat{\v n}_c^\top \m R_{t_{b(j)}} [\v p_j]_\times \\
  \v J_{\m R_{t_{e(j)}}} & = - \alpha_j  \hat{\v n}_c^\top \m R_{t_{e(j)}} [\v p_j]_\times    \\
  \v J_{\v t_{t_{b(j)}}} & = (1-\alpha_j) \hat{\v n}_c^\top                                   \\
  \v J_{\v t_{t_{e(j)}}} & = \alpha_j  \hat{\v n}_c^\top.
\end{align}

The expression $[\cdot]_\times$ is the skew symmetric matrix of a vector.
The derivatives of the error by the transformation parameters of the corresponding point $T_{t_c} = \{\m R_{t_c}, \v t_{t_c}\}$ can be derived analogously
\begin{align}
  \v J_{\m R_{t_{b(c)}}} & = (1-\alpha_c)  \hat{\v n}_c^\top \m R_{t_{b(c)}} [\v p_c]_\times \\
  \v J_{\m R_{t_{e(c)}}} & =  \alpha_c  \hat{\v n}_c^\top \m R_{t_{e(c)}} [\v p_c]_\times    \\
  \v J_{\v t_{t_{b(c)}}} & = -(1-\alpha_c) \hat{\v n}_c^\top                                 \\
  \v J_{\v t_{t_{e(c)}}} & = -\alpha_c  \hat{\v n}_c^\top.
\end{align}
Each error equation given by \eqref{eq:error_function} contributes to the poses from two scans in the optimization.
In the autonomous driving domain, vehicles often operate on large routes where relatively few scans overlap substantially.
This means that the normal equation system is generally sparse. This allows us to utilize sparse matrices and operations to solve the otherwise enormous equation system.

In reality, we do not know which points correspond to each other, nor can we even ensure that a corresponding point was measured.
Therefore, we use the common assumption as in ICP and search for the closest point in a certain neighborhood~$N(\v p_j)$.
We do not only have two point clouds, thus we have to search for correspondences in multiple, and in the worst case, in all the point clouds.
To reduce the compute requirements, we sample a set of scans in which we look for correspondences, as will be further explained in \secref{sec:acceleration}.
Furthermore, we need a sufficiently accurate initial guess required for solving the non-linear least squares problem.
In our experiments, having a LiDAR odometry system combined with loop closures or with a low-cost GPS for global alignment obtained a sufficient initial guess.
To deal with outliers, \eg, caused by misalignment or dynamic objects, we can use a Geman McClure-Kernel to reduce their impact in the optimization.

\subsection{Acceleration Strategies}
\label{sec:acceleration}
Aligning all point clouds can be very time-consuming.
To tackle the problem, we utilize three strategies.
First, instead of matching each point cloud $\mathcal{P}_i$ to all other point clouds, we randomly sample for each point cloud $\mathcal P_i$, $k=1,\dots, N_{\text{matches}}$ different point clouds $\mathcal C_k$ within a radius~$\tau$
\begin{equation}
  \mathcal C_k = \{\mathcal P_k \in \mathcal D \setminus \mathcal P_i \mid \norm{\v t_i -\v t_k} \leq \tau\},
\end{equation}
and only search for point associations in this subset of point clouds.
This allows us to reduce the complexity of the correspondence search from $O(N_\text{Scans}^2)$ to $O(N_\text{Scans})$.

Second, we utilize a voxel hash map similar to Nie{\ss}ner \etalcite{niessner2013siggraph} with bucket size 1.
We store in each voxel the index of the point closest to the voxel center.
Saving the index allows us to also access the point attributes without occupying additional memory.
The voxel hash map construction time is $O(N)$, and performing a radius neighborhood search can be done in $O(1)$.
Note that we have to construct the acceleration structure at each iteration of the optimization, since the local point distribution changes due to the deskewing through the continuous-time trajectory.
Therefore, having an efficient acceleration structure like the voxel hash map is crucial.
Additionally, the hash map can be implemented efficiently on the GPU to facilitate modern hardware accelerators.

Third, we subsample the point clouds.
We utilize a grid-based subsampling, where we can utilize the aforementioned voxel hash map.
Subsampling is a simple but efficient way to save on redundant points that do not add a lot of additional information~\cite{vizzo2023ral,zhang2014rss}.

\label{sec:exp}
\begin{figure}
  \centering
  \begin{subfigure}{0.9\linewidth}
    \includegraphics[width=\textwidth]{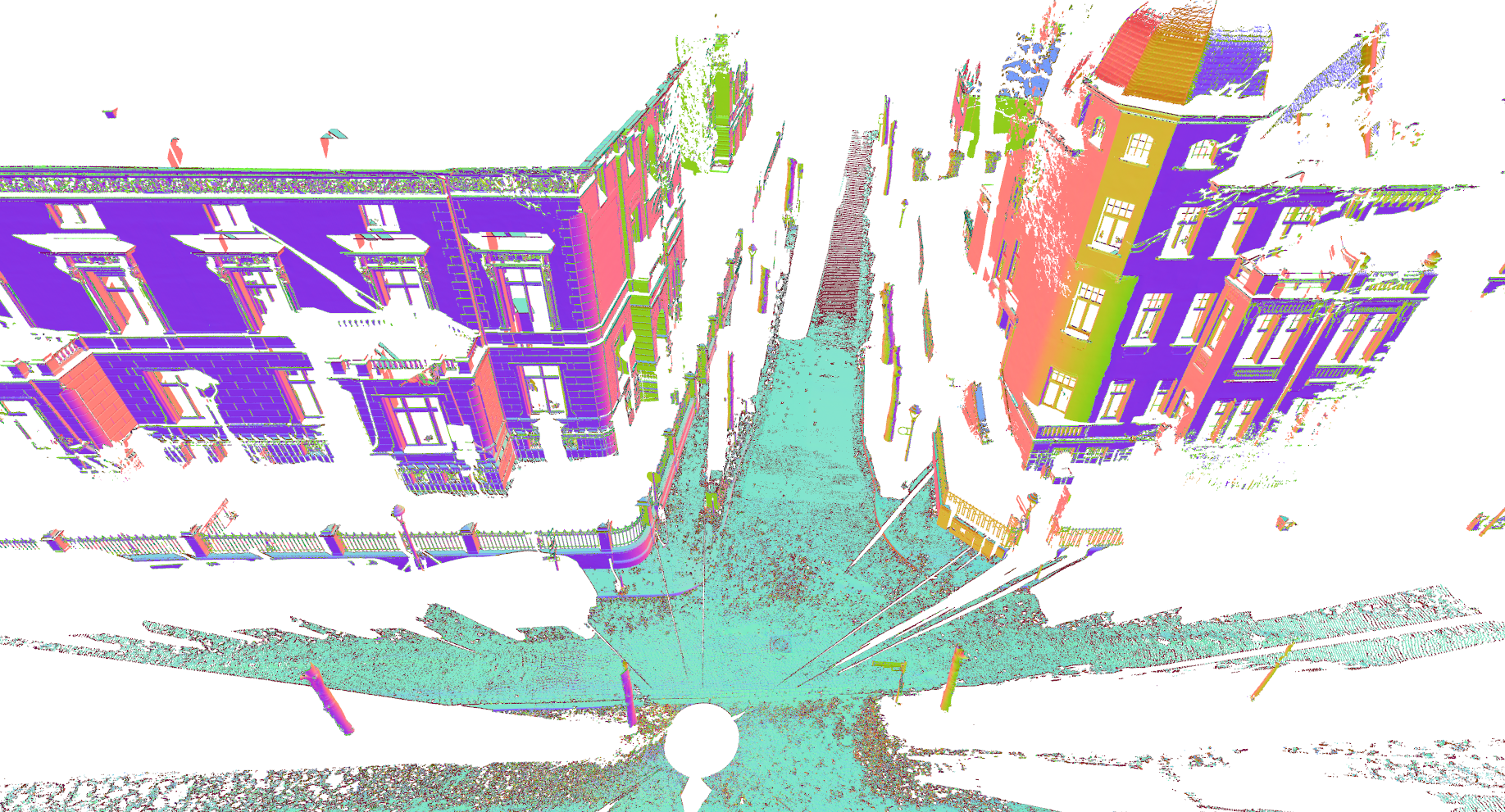}
    \vspace{-0.6cm}
    \caption{TLS scan}
    \vspace{0.1cm}
    \label{fig:tls}
  \end{subfigure}
  \hfill
  \begin{subfigure}{0.9\linewidth}
    \includegraphics[width=\textwidth]{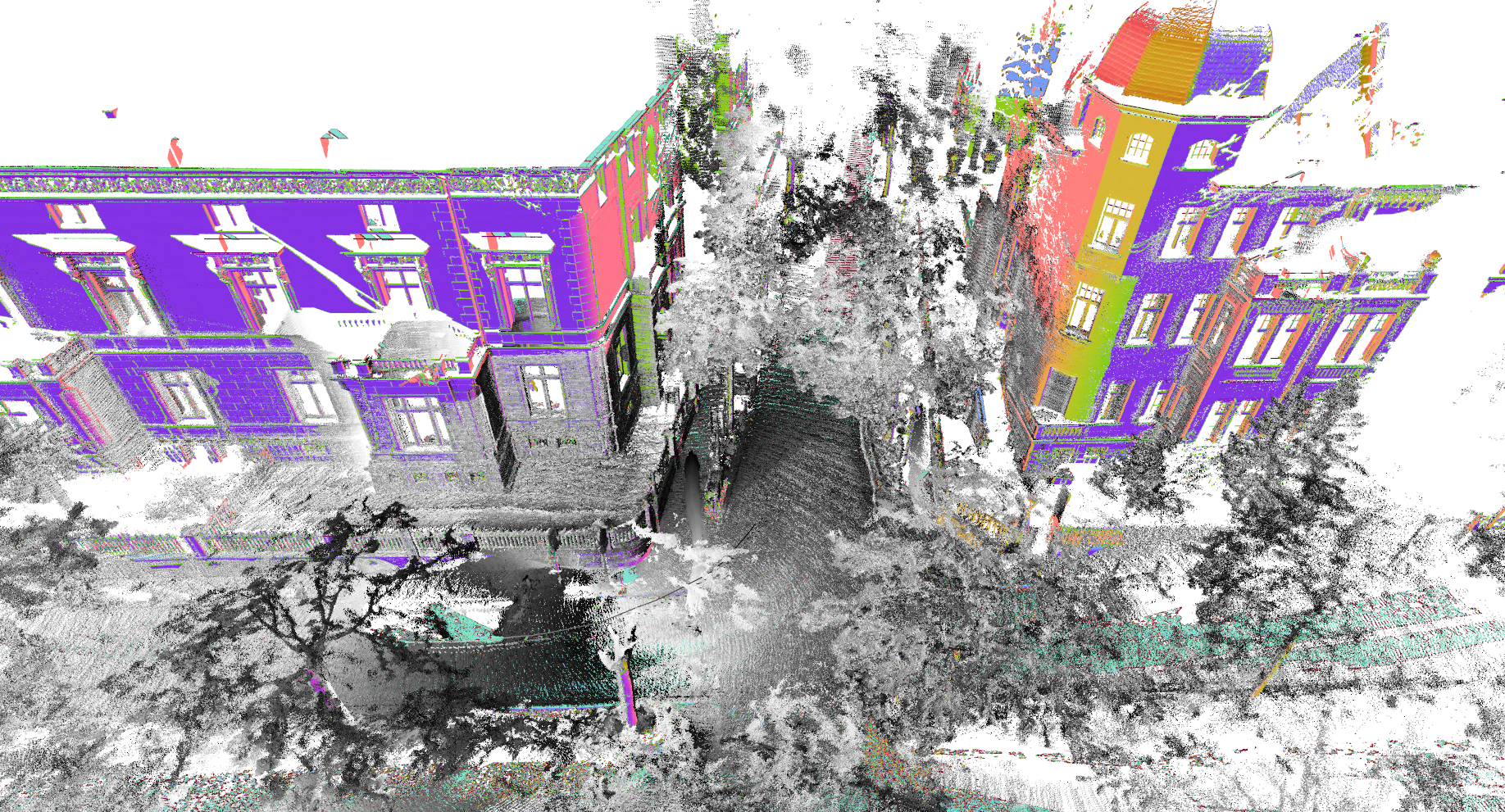}
    \vspace{-0.6cm}
    \caption{LiDAR Scans aligned to TLS scan}
    \vspace{0.1cm}
    \label{fig:tls_aligned}
  \end{subfigure}

  \caption{(a) The ground truth for our dataset was generated using high-precision, globally referenced TLS scans. (b) Close-by LiDAR scans (gray) are aligned to the TLS scans to obtain reference poses. Note that we removed moving objects (cars, pedestrians) and movable objects (parked cars, foliage) manually from the TLS to allow precise scan alignment.  }
  \label{fig:tls_gt}
\end{figure}
\subsection{Memory Management}
Storing all point clouds in memory is infeasible due to the sheer amount of data when considering realistic setups.
Loading each point cloud on demand from disk is memory-efficient but slow, since reading from RAM is usually way faster than from disk.
We compromise by using a circular buffer, only keeping the last $N_\text{buffer}$ used point clouds in RAM.
Whenever a point cloud is required that is not in the buffer, we delete the scan from the buffer (but still keep it on disk) that was used last and replace it with the new scan.
Utilizing such out-of-core processing allows for processing large amounts of data efficiently, without relying on large RAM resources.
Due to sampling the point clouds based on the distance $\tau$, it is quite likely to sample point clouds that are also close in time.
Therefore, we iterate sequentially through the point clouds to reuse, as much as possible, the point clouds in the buffer before loading new ones in.
Additionally, we reverse after each optimization step the iteration order of the point clouds, such that the end of the last iteration is now the beginning of the new one for additional speed-up.
\section{Experimental Evaluation}

The main focus of this work is LiDAR bundle adjustment, where we want to optimize from a set of scans and initial poses the continuous-time trajectory that globally aligns the point clouds with respect to each other.
We present our experiments to show the capabilities of our method. The results of our experiments also support our key claims, which are:
(i)~Our approach is able to efficiently estimate a precise continuous-time trajectory,
(ii)~provides accurate, globally aligned point cloud maps, and
(iii)~can perform single- and multi-session alignment.

In the following experiments, we precompute the normals of the point clouds using its 30 nearest neighbors in the scan. For each point cloud, we search in $N_\text{matches}=10$ other clouds within $\tau=30$\,m radius for matches. We downsample the scans to a resolution of 15\,cm while we use for the voxel hash-map a grid size of 30\,cm for finding the correspondences. We only look for the correspondences in the adjacent voxels, resulting in $3^3=27$ candidates. Furthermore, we use a buffer size $N_\text{buffer}=1000$ point clouds, which practically results in the same runtime as loading all point clouds in memory. For the non-linear least squares adjustment, we stop at the latest after $N_\text{iter}=100$ iterations. Usually, after around 10-30 iterations (depending on the quality of the initial alignment), the adjustment converges. All experiments were conducted with an NVIDIA RTX-A5000.

\subsection{Results on City-scale Data}

\begin{table}[t]
  \caption{Quantiative results on our recorded IPB-Car Dataset}
  \label{tab:ipb}
  \begin{tabularx}{\linewidth}{c||C|C|C|C}
    \toprule
    Approach       & ATE [m] (trans)                               & ATE [$^\circ$] (rot)                          & RPE [m] (trans) & RPE [$^\circ$] (rot) \\ 
    \midrule
    KISS-ICP       &  2.28 & 1.21
     &  0.023 & 0.121
                                              \\
    KISS-ICP + GPS &  1.34 & 2.79
 &  0.025 & 0.131
                                          \\
    PIN SLAM       &  1.76 & 1.43
 &  0.020 & 0.122
                                          \\
    PBA            &  0.99 & 3.53 
    & N/A                                           & N/A                                    \\
    HBA            &  1.29 & 2.55
      &  0.026 & 0.133
                                               \\
    Our approach   & \textbf{ 0.90} & \textbf{0.63}
      & \textbf{ 0.014} & \textbf{0.055}
                                               \\
    \bottomrule
  \end{tabularx}
\end{table}
\begin{figure}[t]
  \centering
  \includegraphics[width=0.9\linewidth]{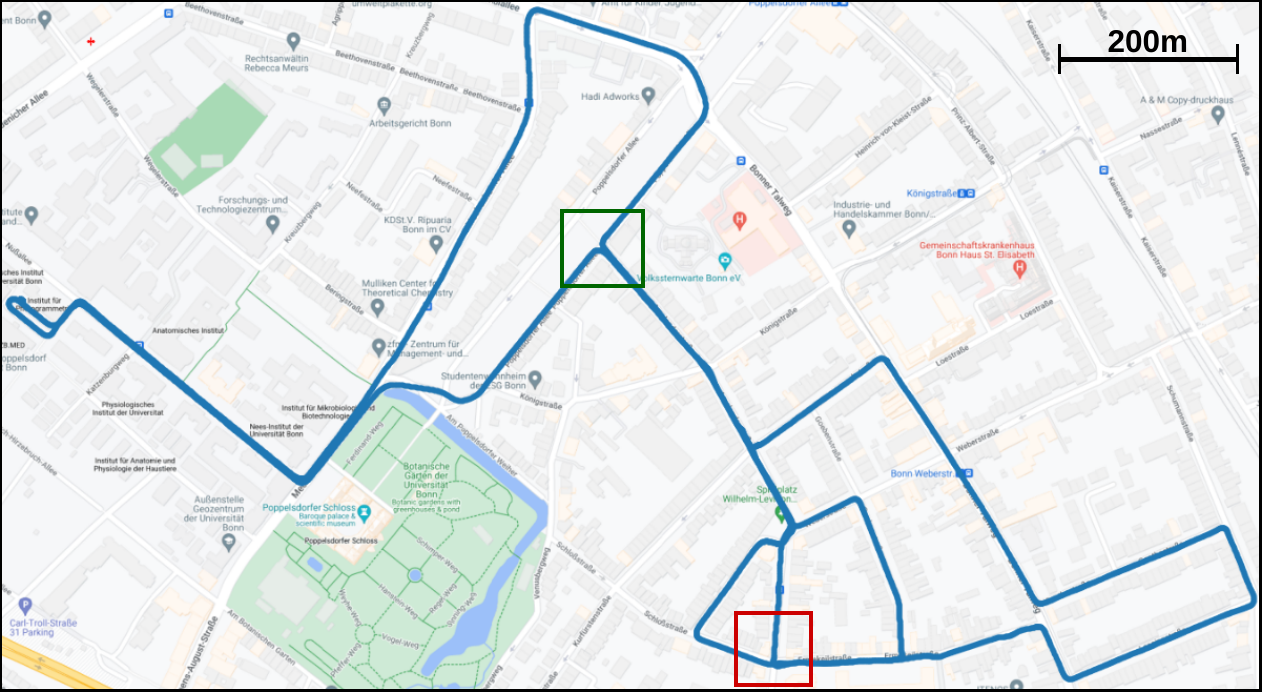}
  \caption{The estimated trajectory of our approach on the self-recorded IPB-Car dataset. The trajectory contains multiple loop closures, where some sections are mapped multiple times. Making a global alignment necessary to obtain a consistent trajectory.
  }
  \label{fig:map}
\end{figure}
\begin{figure*}[t]
  \centering
  \includegraphics[width=0.95\linewidth]{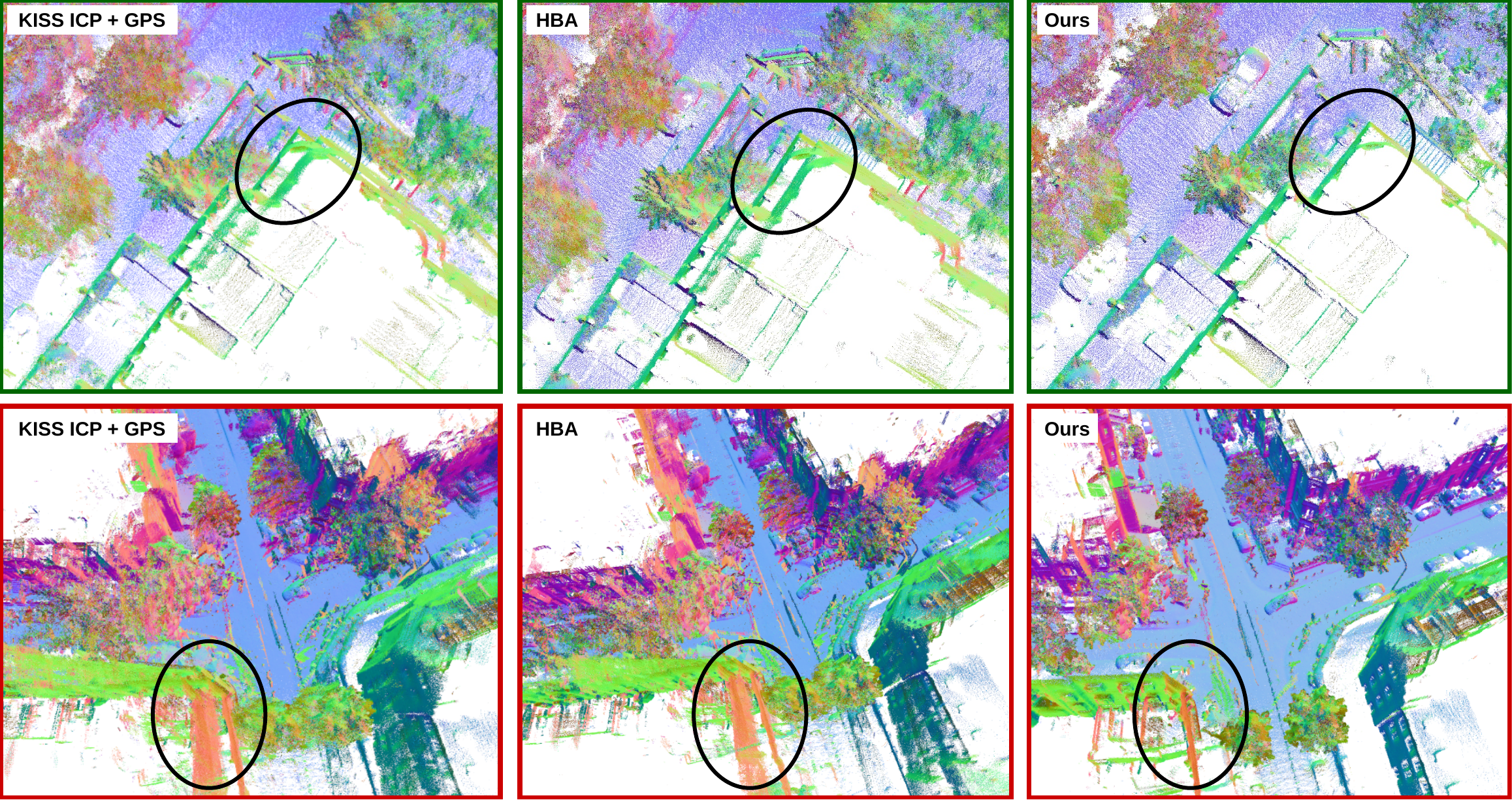}
  \caption{Shown are two close-up parts of the IPB-car dataset to allow qualitative evaluation. The initial guess for the methods was obtained using KISS-ICP fused with GPS data (left). Due to bad GPS conditions, the map is misaligned after closing longer loops, resulting in an inconsistent map. HBA can only slightly improve the local map quality, while our approach is able to obtain a crisp point cloud map.  }
  \label{fig:qualitative}
\end{figure*}

The first experiment evaluates the performance of our approach on our own collected dataset, which we will refer from here on as the IPB-Car dataset.
Our setup consists of a car mount equipped with multiple sensors.
We use a horizontally mounted Ouster1-128 that provides us with point clouds recorded with 128 laser beams at 10\,Hz, and a low-cost GPS receiver to provide global reference.
The dataset contains 11702 scans with each ${N=128 \cdot 2048}$ points.
The initial guess for our approach is obtained by fusing the poses from KISS-ICP~\cite{vizzo2023ral} as odometry and GPS as unity factors in a pose graph.

For evaluation, a ground truth for more accurate reference data is needed. For that, we selected key locations in the map, which we measured with a terrestrial laser scanner (TLS). The TLS scans are globally referenced using precise static dual frequency GPS with long static recording per reference scan and using correction data from SAPOS to mitigate atmospheric errors, 
resulting in centimeter-accurate global reference poses with millimeter accurate local accuracy.
We align the recorded LiDAR scans to the TLS ground truth scans similar to Nguyen~\etalcite{nguyen2024cvpr} for precise reference poses.
In this dataset, the vehicle passed 8 times through reference locations.
All vehicle LiDAR scans that are aligned to a TLS scan will be used for quantitative evaluation.
As metrics, we will use the common absolute trajectory error (ATE) to obtain a measure for the global positioning accuracy, as well as the relative position error (RPE) as a measure for local accuracy.

We also compare our approach against state-of-the-art baselines, here specifically against PIN-SLAM~\cite{pan2024tro} an online SLAM approach, as well as the LiDAR bundle adjustment methods HBA~\cite{liu2023ral} and PBA~\cite{giammarino2023ral}. For completeness, we provide the results of KISS-ICP~\cite{vizzo2023ral} solely as odometry system and in a second variant fused with GPS measurements as initial guess.
The results of our collected dataset are shown in \tabref{tab:ipb}.
Our approach is able to consistently outperform all baselines regarding ATE, as well as RPE.
Our approach takes on average 45\,min per iteration for the 11702 scans, resulting in around two days of overall optimization time.
Note that PBA~\cite{giammarino2023ral} only estimates the poses for keyframes, which does not allow us to compute the RPE, since the metric is computed between consecutive poses.
HBA~\cite{liu2023ral} requires around 160\,GB of CPU memory, while PBA~\cite{giammarino2023ral} requires 43\,GB GPU memory even though downsampling the poses.
With our out-of-core ring buffer containing the last 1000 scans, we only require 8\,GB of GPU memory.
Note that one can further reduce the memory consumption by reducing the buffer size, but trading off slower processing time.
Just storing naively the full normal equation system without exploiting the sparsity would alone require 40\,GB memory, while solving would take over 150 days.

\subsection{Qualitative Results}
To convey the quality of the estimated trajectory, we will show the resulting aggregated maps.
In \figref{fig:map} we show our estimated trajectory (blue), as well as a red and green rectangle, indicating the scenes shown in \figref{fig:qualitative}.
We visualize the aggregated point cloud map using the estimated poses of the approaches, where the color is coded based on the normals.
The green area is driving through a parkway with a lot of trees, resulting in bad GPS conditions.
This leads to a translation error in the initial guess (KISS-ICP + GPS), as can be seen by the walls not being aligned.
Starting from this, HBA is not able to resolve the local inconsistencies, while our approach successfully aligns the point clouds.

The area denoted by the red frame is in an urban environment with high, terraced houses, making it susceptible to GPS multipathing.
Additionally, misalignment in a sharp turn caused blurry maps. HBA can only reduce the blurring slightly, while our approach is able to correct the pose errors and provide sharp walls.

\subsection{Results on Campus-scale Data}
\begin{table}[t]
  \caption{Quantiative results on the MCD~\cite{nguyen2024cvpr} dataset}
  \label{tab:mcd}
  \begin{tabularx}{\linewidth}{cc||C|C|C|C}
    \toprule
     & Approach        & ATE (trans)                                     & ATE (rot)                                       & RPE (trans) & RPE (rot) \\ \midrule
    \multirow{5}{*}{\rotatebox[origin=c]{90}{NTU-day-1}}
     & KISS-ICP        &  7.21 & 3.54
      &  0.133 & 1.056
                                \\
     & KISS-ICP + Loop &  2.08 & 3.66
 &  0.129 & 1.059
                           \\
     & PIN SLAM        &  1.34 & 2.20
  & \textbf{ 0.086} & 1.069
                            \\
     & HBA             &  1.79 & 2.81
       &  0.129 & 1.060
                                 \\
     & Our approach    & \textbf{ 1.03} & \textbf{1.75}
       &  0.129 & \textbf{0.599}
                                 \\
    \midrule
    \multirow{5}{*}{\rotatebox[origin=c]{90}{NTU-day-2}}
     & KISS-ICP        &  0.27 & \textbf{0.86}
      &  0.094 & 0.663
                                \\
     & KISS-ICP + Loop &  0.35 & 1.64
 &  0.092 & 0.665
                           \\
     & PIN SLAM        &  0.28 & 1.29
  & \textbf{ 0.063} & 0.636
                            \\
     & HBA             & \textbf{ 0.19} & 0.87
       &  0.092 & 0.665
                                 \\
     & Our approach    &  0.20 & 0.99
       &  0.083 & \textbf{0.397}
                                 \\
    \bottomrule
  \end{tabularx}
\end{table}

To validate our performance in a different environment under different conditions, we evaluate on the Multi-Campus Dataset (MCD)~\cite{nguyen2024cvpr} dataset.
We choose this dataset due to the high-quality ground truth that they also created using precise TLS reference maps.
Since our approach refines poses from, \eg, SLAM/odometry + GPS, comparing against ground truth generated with similar methods~\cite{geiger2012cvpr,weixin2019cvpr} would not be significant.
In contrast to our dataset, where the sensor is mounted on a car and the environment is in an urban environment, MCD~\cite{nguyen2024cvpr} uses a handheld device in a campus environment.
The ntu-day1 sequence has 6010 scans, while ntu-day2 is with 2274 scans substantially smaller.
Due to a lack of GPS reception, we use KISS-ICP~\cite{vizzo2023ral} optimized with loop closures~\cite{gupta2024icra} using pose graph optimization as an initial guess.
The results are depicted in \tabref{tab:mcd}.
On the longer day1 sequence, our approach is able to estimate the most accurate trajectory in terms of ATE (translation and rotation), and RPE (rotation), as well as the second best regarding RPE (translation).
On the shorter sequence, the quality of the estimated trajectories is more similar. Our approach provides throughout competitive results.

\subsection{Mutli-Session Alignment}
\begin{table}[t]
  \caption{Multi-Session Alignment MCD-NTU day 1 \& 2}
  \label{tab:multisession}
  \begin{tabularx}{\linewidth}{c||C|C}
    \toprule
    Approach        & Inter RPE [m] (trans)                                & Inter RPE [$^\circ$] (rot) \\ \midrule
    KISS-ICP + Loop &  0.567 & 3.20 
                              \\
    Our             & \textbf{ 0.085} & \textbf{0.08}
                                    \\
    \bottomrule
  \end{tabularx}
\end{table}

\begin{figure}
  \centering
  \begin{subfigure}{0.85\linewidth}
    \includegraphics[width=\textwidth]{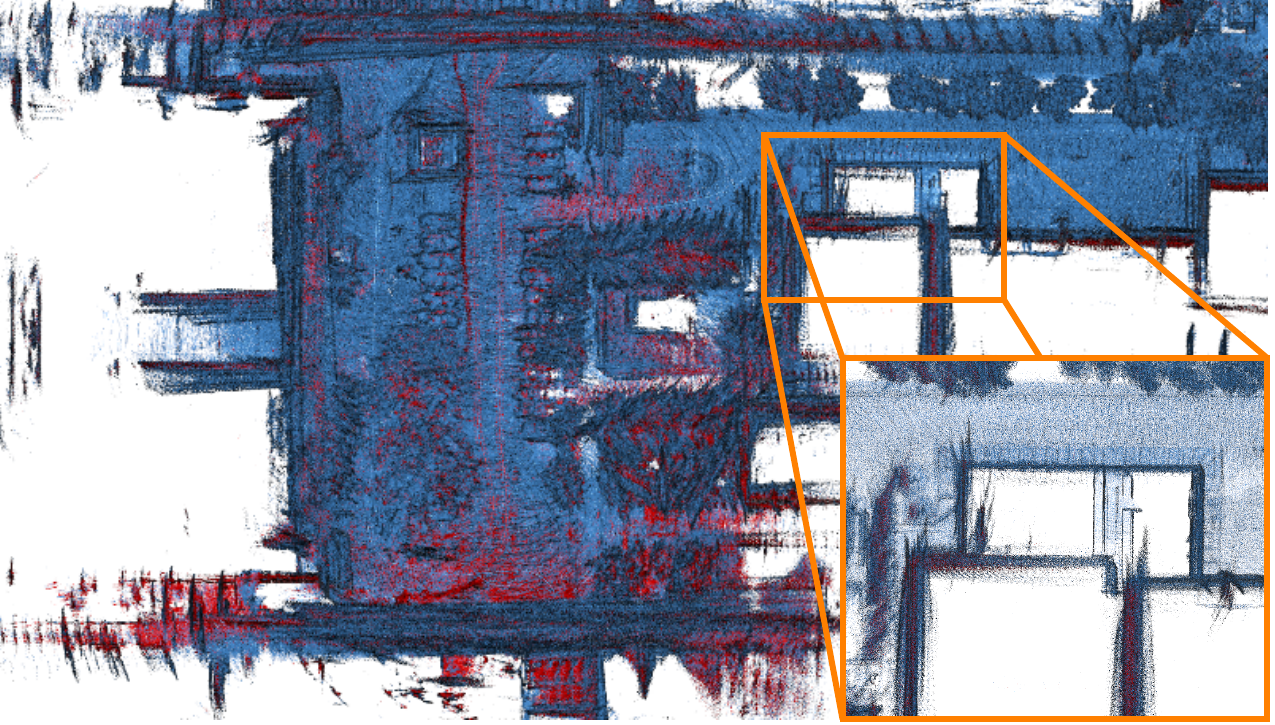}
    \vspace{-0.6cm}
    \caption{Initial guess} 
    \vspace{0.1cm}
    \label{fig:multi_kiss}
  \end{subfigure}
  \hfill
  \begin{subfigure}{0.85\linewidth}
    \includegraphics[width=\textwidth]{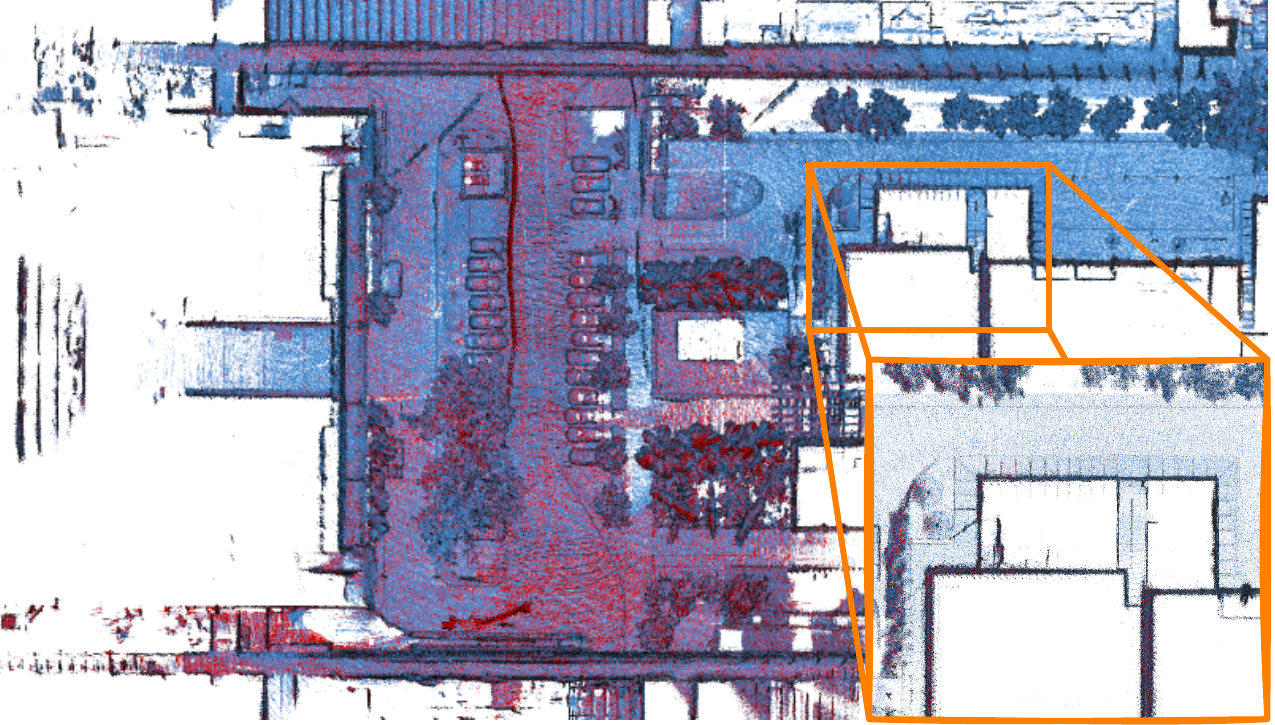}
    \vspace{-0.6cm}
    \caption{Estimated alignment}
    \vspace{0.1cm}
    \label{fig:multi_ours}
  \end{subfigure}

  \caption{For the multi-session alignment, we jointly optimize the scans from two sessions (Session 1: Blue, Session 2: red). (a) Depicts the initial guess, where one can see that the sessions are roughly aligned, but overall the map is quite noisy. In (b), we show the aggregated map after our alignment. Seeing that both the red and blue point clouds are well aligned; leading to crisp structures.  }
  \label{fig:multisession}
\end{figure}

The MCD~\cite{nguyen2024cvpr} dataset provides different runs in the same location on different days, enabling us to perform and test multi-session alignment.
In this task, we want to align multiple scans from multiple sessions together, such that they form a consistent map.
This task is important when building maps of areas, that cannot be recorded in one session, or for comparing/ combining data recorded at different points in time.
For the initial guess, we take the previously separately estimated KISS-ICP~\cite{vizzo2023ral} poses and fuse it with loop closures w.r.t the first sequence.
In other words, combining LiDAR odometry, plus inter-loop closure within the first sequence and intra-loop closure from the second sequence to the first sequence in a pose graph.
To run multi-session alignment with our method, we can simply optimize both trajectories jointly.
The only thing one needs to deal with is the time discontinuity between the sessions, \ie, for the continuous-time trajectory, the end pose of the first session is \emph{not} the start pose of the second session.

For quantitative evaluation,
we search for each pose of the second session the closest pose of the first session.
For each of those corresponding pairs, we compute the relative transformation.
We do this once for the ground truth poses and also for the estimated ones (using for both the same pose correspondences).
Between both sets of relative transformations, we compute the root mean squared error. This is similar to the RPE but between two different sessions, therefore we will denote it as inter-RPE.
The results for the initial guess and for our approach are provided in \tabref{tab:multisession}.
We see that our approach is able to reduce the initial alignment error substantially.
For qualitative comparison, we refer to \figref{fig:multisession}.
Although the trajectories of the initial guess are globally well aligned, the resulting aggregated map is noisy, especially due to errors in rotation.
Our approach however is able to align the sessions, resulting in a joint, globally and locally aligned map.

\section{Conclusion}
\label{sec:conclusion}
In this paper, we presented a novel approach for LiDAR bundle adjustment.
Our approach takes a large set of scans and estimates a refined continuous-time trajectory, resulting in a globally aligned point cloud map.
We jointly align the set of overlapping scans by using a robust least squares formulation.
Reducing the search for potential correspondences between the scans, as well as utilizing hash maps, allows for globally aligning thousands of scans in a reasonable time.
The usage of a circular buffers allows for scaling up the sequence length without running into memory problems.
We implemented and evaluated our approach on different datasets with TLS-based, accurate ground truth information.
This allows us to successfully estimate the trajectory of the 3D LiDAR sensor and provide well-aligned point cloud maps.
Our approach is able to provide competitive results on different datasets.
We show that our approach can provide accurate trajectories, tested on sequences with up to 11700 scans.
Additionally, we show that our approach can be used to jointly align multiple sessions.

\section*{Acknowledgments}
We greatly thank Martin Blome for measuring the reference locations with the TLS and the global alignment.

\bibliographystyle{plain_abbrv}

\bibliography{glorified,new}

\begin{thebibliography}{10}

\bibitem{behley2018rss}
J.~Behley and C.~Stachniss.
\newblock {Efficient Surfel-Based SLAM using 3D Laser Range Data in Urban
  Environments}.
\newblock In {\em Proc.~of Robotics: Science and Systems (RSS)}, 2018.

\bibitem{benjemaa1998eccv}
R.~Benjemaa and F.~Schmitt.
\newblock A solution for the registration of multiple 3d point sets using unit
  quaternions.
\newblock In {\em Proc.~of the Europ.~Conf.~on Computer Vision (ECCV)}, 1998.

\bibitem{bergevin1996tpami}
R.~Bergevin, M.~Soucy, H.~Gagnon, and D.~Laurendeau.
\newblock Towards a general multi-view registration technique.
\newblock {\em IEEE Trans.~on Pattern Analysis and Machine Intelligence
  (TPAMI)}, 18(5):540--547, 1996.

\bibitem{besl1992pami}
P.~Besl and N.~McKay.
\newblock {A Method for Registration of 3D Shapes}.
\newblock {\em IEEE Trans.~on Pattern Analysis and Machine Intelligence
  (TPAMI)}, 14(2):239--256, 1992.

\bibitem{biber2003iros}
P.~Biber and W.~Stra{\ss}er.
\newblock The normal distributions transform: A new approach to laser scan
  matching.
\newblock In {\em Proc.~of the IEEE/RSJ Intl.~Conf.~on Intelligent Robots and
  Systems (IROS)}, 2003.

\bibitem{chen1991iros}
Y.~Chen and G.~Medioni.
\newblock {Object Modelling by Registration of Multiple Range Images}.
\newblock In {\em Proc.~of the IEEE/RSJ Intl.~Conf.~on Intelligent Robots and
  Systems (IROS)}, 1991.

\bibitem{dai2019rs}
J.~Dai, L.~Yan, H.~Liu, C.~Chen, and L.~Huo.
\newblock An offline coarse-to-fine precision optimization algorithm for 3d
  laser slam point cloud.
\newblock {\em Remote Sensing}, 11(20):2352, 2019.

\bibitem{das2013icra-3sru}
A.~Das, J.~Servos, and S.~Waslander.
\newblock {3D Scan Registration Using the Normal Distributions Transform with
  Ground Segmentation and Point Cloud Clustering}.
\newblock In {\em Proc.~of the IEEE Intl.~Conf.~on Robotics \& Automation
  (ICRA)}, 2013.

\bibitem{dellenbach2022icra}
P.~Dellenbach, J.~Deschaud, B.~Jacquet, and F.~Goulette.
\newblock {CT-ICP Real-Time Elastic LiDAR Odometry with Loop Closure}.
\newblock In {\em Proc.~of the IEEE Intl.~Conf.~on Robotics \& Automation
  (ICRA)}, 2022.

\bibitem{deschaud2018icra}
J.~Deschaud.
\newblock {IMLS-SLAM: scan-to-model matching based on 3D data}.
\newblock In {\em Proc.~of the IEEE Intl.~Conf.~on Robotics \& Automation
  (ICRA)}, 2018.

\bibitem{giammarino2023ral}
L.~Di~Giammarino, E.~Giacomini, L.~Brizi, O.~Salem, and G.~Grisetti.
\newblock {Photometric LiDAR and RGB-D Bundle Adjustment}.
\newblock {\em IEEE Robotics and Automation Letters (RA-L)}, 8(7):4362--4369,
  2023.

\bibitem{droeschel2018icra}
D.~Droeschel and S.~Behnke.
\newblock {Efficient Continuous-Time SLAM for 3D Lidar-Based Online Mapping}.
\newblock In {\em Proc.~of the IEEE Intl.~Conf.~on Robotics \& Automation
  (ICRA)}, 2018.

\bibitem{einhorn2015ras}
E.~Einhorn and H.M. Gross.
\newblock Generic ndt mapping in dynamic environments and its application for
  lifelong slam.
\newblock {\em Journal on Robotics and Autonomous Systems (RAS)}, 69:28--39,
  2015.

\bibitem{geiger2012cvpr}
A.~Geiger, P.~Lenz, and R.~Urtasun.
\newblock {Are we ready for Autonomous Driving? The KITTI Vision Benchmark
  Suite}.
\newblock In {\em Proc.~of the IEEE Conf.~on Computer Vision and Pattern
  Recognition (CVPR)}, 2012.

\bibitem{gelfand2005robust}
N.~Gelfand, N.J. Mitra, L.J. Guibas, and H.~Pottmann.
\newblock {Robust Global Registration}.
\newblock In {\em Proc.~of the Symp. on Geometry Processing}, 2005.

\bibitem{gupta2024icra}
S.~Gupta, T.~Guadagnino, B.~Mersch, I.~Vizzo, and C.~Stachniss.
\newblock {Effectively Detecting Loop Closures using Point Cloud Density Maps}.
\newblock In {\em Proc.~of the IEEE Intl.~Conf.~on Robotics \& Automation
  (ICRA)}, 2024.

\bibitem{johnson1999ivc}
A.E. Johnson and S.B. Kang.
\newblock {Registration and Integration of Textured 3D Data}.
\newblock {\em Journal on Image and Vision Computing (IVC)}, 17(2):135--147,
  1999.

\bibitem{kaess2008tro}
M.~Kaess, A.~Ranganathan, and F.~Dellaert.
\newblock {iSAM}: Incremental smoothing and mapping.
\newblock {\em IEEE Trans.~on Robotics (TRO)}, 24(6):1365--1378, 2008.

\bibitem{keller2013threedv}
M.~Keller, D.~Lefloch, M.~Lambers, and S.~Izadi.
\newblock {Real-time 3D Reconstruction in Dynamic Scenes using Point-based
  Fusion}.
\newblock In {\em Proc.~of the Intl.~Conf.~on 3D Vision (3DV)}, 2013.

\bibitem{kerl2013icra}
C.~Kerl, J.~Sturm, and D.~Cremers.
\newblock {Robust Odometry Estimation for RGB-D Cameras}.
\newblock In {\em Proc.~of the IEEE Intl.~Conf.~on Robotics \& Automation
  (ICRA)}, 2013.

\bibitem{li2024arxiv}
J.~Li, T.M. Nguyen, S.~Yuan, and L.~Xie.
\newblock Pss-ba: Lidar bundle adjustment with progressive spatial smoothing.
\newblock {\em arXiv preprint}, arXiv:2403.06124, 2024.

\bibitem{liu2023ral}
X.~Liu, Z.~Liu, F.~Kong, and F.~Zhang.
\newblock Large-scale lidar consistent mapping using hierarchical lidar bundle
  adjustment.
\newblock {\em IEEE Robotics and Automation Letters (RA-L)}, 8(3):1523--1530,
  2023.

\bibitem{liu2021ral}
Z.~Liu and F.~Zhang.
\newblock Balm: Bundle adjustment for lidar mapping.
\newblock {\em IEEE Robotics and Automation Letters (RA-L)}, 6(2):3184--3191,
  2021.

\bibitem{weixin2019cvpr}
W.~Lu, Y.~Zhou, G.~Wan, S.~Hou, and S.~Song.
\newblock L3-net: Towards learning based lidar localization for autonomous
  driving.
\newblock In {\em Proc.~of the IEEE/CVF Conf.~on Computer Vision and Pattern
  Recognition (CVPR)}, 2019.

\bibitem{newcombe2011ismar}
R.A. Newcombe, S.~Izadi, O.~Hilliges, D.~Molyneaux, D.~Kim, A.J. Davison,
  P.~Kohli, J.~Shotton, S.~Hodges, and A.~Fitzgibbon.
\newblock {KinectFusion: Real-Time Dense Surface Mapping and Tracking}.
\newblock In {\em Proc.~of the Intl.~Symp.~on Mixed and Augmented Reality
  (ISMAR)}, 2011.

\bibitem{nguyen2024cvpr}
T.M. Nguyen, S.~Yuan, T.H. Nguyen, P.~Yin, H.~Cao, L.~Xie, M.~Wozniak,
  P.~Jensfelt, M.~Thiel, J.~Ziegenbein, and N.~Blunder.
\newblock {MCD: Diverse Large-Scale Multi-Campus Dataset for Robot Perception}.
\newblock In {\em Proc.~of the IEEE/CVF Conf.~on Computer Vision and Pattern
  Recognition (CVPR)}, 2024.

\bibitem{niessner2013siggraph}
M.~Nie{\ss}ner, M.~Zollh{\"o}fer, S.~Izadi, and M.~Stamminger.
\newblock {Real-time 3D Reconstruction at Scale using Voxel Hashing}.
\newblock In {\em Proc.~of the SIGGRAPH Asia}, 2013.

\bibitem{pan2024tro}
Y.~Pan, X.~Zhong, L.~Wiesmann, T.~Posewsky, J.~Behley, and C.~Stachniss.
\newblock Pin-slam: Lidar slam using a point-based implicit neural
  representation for achieving global map consistency.
\newblock {\em IEEE Trans.~on Robotics (TRO)}, 2024.

\bibitem{park2018icra-elfd}
C.~Park, P.~Moghadam, S.~Kim, A.~Elfes, C.~Fookes, and S.~Sridharan.
\newblock {Elastic LiDAR Fusion: Dense Map-Centric Continuous-Time SLAM}.
\newblock In {\em Proc.~of the IEEE Intl.~Conf.~on Robotics \& Automation
  (ICRA)}, 2018.

\bibitem{park2022tro}
C.~Park, P.~Moghadam, J.L. Williams, S.~Kim, S.~Sridharan, and C.~Fookes.
\newblock {Elasticity Meets Continuous-Time: Map-Centric Dense 3D LiDAR SLAM}.
\newblock {\em IEEE Trans.~on Robotics (TRO)}, 38(2):978--997, 2022.

\bibitem{park2017iccv}
J.~Park, Q.~Zhou, and V.~Koltun.
\newblock {Colored Point Cloud Registration Revisited}.
\newblock In {\em Proc.~of the IEEE Intl.~Conf.~on Computer Vision (ICCV)},
  2017.

\bibitem{rusinkiewicz2001dim}
S.~Rusinkiewicz and M.~Levoy.
\newblock {Efficient variants of the ICP algorithm}.
\newblock In {\em Proc.~of Intl.~Conf.~on 3-D Digital Imaging and Modeling},
  2001.

\bibitem{saarinen2013iros-f3mi}
J.~Saarinen, T.~Stoyanov, H.~Andreasson, and A.~Lilienthal.
\newblock {Fast 3D Mapping in Highly Dynamic Environments Using Normal
  Distributions Transform Occupancy Maps}.
\newblock In {\em Proc.~of the IEEE/RSJ Intl.~Conf.~on Intelligent Robots and
  Systems (IROS)}, 2013.

\bibitem{schneider2012isprs}
J.~Schneider, F.~Schindler, T.~L\"abe, and W.~F\"orstner.
\newblock Bundle adjustment for multi-camera systems with points at infinity.
\newblock In {\em ISPRS Annals of Photogrammetry, Remote Sensing and Spatial
  Information Sciences}, 2012.

\bibitem{segal2009rss}
A.~Segal, D.~Haehnel, and S.~Thrun.
\newblock {Generalized-ICP}.
\newblock In {\em Proc.~of Robotics: Science and Systems (RSS)}, 2009.

\bibitem{shan2020iros}
T.~Shan, B.~Englot, D.~Meyers, W.~Wang, C.~Ratti, and D.~Rus.
\newblock {LIO-SAM: Tightly-coupled Lidar Inertial Odometry via Smoothing and
  Mapping}.
\newblock In {\em Proc.~of the IEEE/RSJ Intl.~Conf.~on Intelligent Robots and
  Systems (IROS)}, 2020.

\bibitem{shan2018iros}
T.~Shan and B.~Englot.
\newblock {LeGO-LOAM: Lightweight and Ground-Optimized Lidar Odometry and
  Mapping on Variable Terrain}.
\newblock In {\em Proc.~of the IEEE/RSJ Intl.~Conf.~on Intelligent Robots and
  Systems (IROS)}, 2018.

\bibitem{skuddis2024arxiv}
D.~Skuddis and N.~Haala.
\newblock Dmsa--dense multi scan adjustment for lidar inertial odometry and
  global optimization.
\newblock {\em arXiv preprint}, arXiv:2402.19044, 2024.

\bibitem{steinbruecker2014icra}
F.~Steinbr{\"u}cker, J.~Sturm, and D.~Cremers.
\newblock {Volumetric 3D Mapping in Real-Time on a CPU}.
\newblock In {\em Proc.~of the IEEE Intl.~Conf.~on Robotics \& Automation
  (ICRA)}, 2014.

\bibitem{stoyanov2012ijrr}
T.~Stoyanov, M.~Magnusson, H.~Andreasson, and A.J. Lilienthal.
\newblock {Fast and accurate scan registration through minimization of the
  distance between compact 3D NDT representations}.
\newblock {\em Intl.~Journal~of Robotics Research (IJRR)}, 31(12):1377--1393,
  2012.

\bibitem{stueckler2014vcir}
J.~St{\"u}ckler and S.~Behnke.
\newblock {Multi-Resolution Surfel Maps for Efficient Dense 3D Modeling and
  Tracking}.
\newblock {\em Journal of Visual Communication and Image
  Representation~(JVCIR)}, 25(1):137--147, 2014.

\bibitem{thrun2006ijrr}
S.~Thrun and M.~Montemerlo.
\newblock {The graph SLAM algorithm with applications to large-scale mapping of
  urban structures}.
\newblock {\em Intl.~Journal~of Robotics Research (IJRR)}, 25(5-6):403, 2006.

\bibitem{triggs1999iccv}
B.~Triggs, P.F. McLauchlan, R.I. Hartley, and A.W. Fitzgibbon.
\newblock Bundle adjustment - a modern synthesis.
\newblock In {\em Proc. of the Intl.~Workshop on Vision Algorithms: Theory and
  Practice}, 1999.

\bibitem{vizzo2023ral}
I.~Vizzo, T.~Guadagnino, B.~Mersch, L.~Wiesmann, J.~Behley, and C.~Stachniss.
\newblock {KISS-ICP: In Defense of Point-to-Point ICP -- Simple, Accurate, and
  Robust Registration If Done the Right Way}.
\newblock {\em IEEE Robotics and Automation Letters (RA-L)}, 8(2):1029--1036,
  2023.

\bibitem{wang2021iros-fflo}
H.~Wang, C.~Wang, C.~Chen, and L.~Xie.
\newblock {F-LOAM: Fast LiDAR Odometry and Mapping}.
\newblock In {\em Proc.~of the IEEE/RSJ Intl.~Conf.~on Intelligent Robots and
  Systems (IROS)}, 2021.

\bibitem{wang2021icra}
Y.~Wang, N.~Funk, M.~Ramezani, S.~Papatheodorou, M.~Popovic, M.~Camurri,
  S.~Leutenegger, and M.~Fallon.
\newblock {Elastic and Efficient LiDAR Reconstruction for Large-Scale
  Exploration Tasks}.
\newblock In {\em Proc.~of the IEEE Intl.~Conf.~on Robotics \& Automation
  (ICRA)}, 2021.

\bibitem{whelan2014ijrr}
T.~Whelan, M.~Kaess, H.~Johannsson, M.~Fallon, J.J. Leonard, and J.~McDonald.
\newblock {Real-time large scale dense RGB-D SLAM with volumetric fusion}.
\newblock {\em Intl.~Journal~of Robotics Research (IJRR)}, 34(4-5):598--626,
  2014.

\bibitem{whelan2015rss}
T.~Whelan, S.~Leutenegger, R.S. Moreno, B.~Glocker, and A.~Davison.
\newblock {ElasticFusion: Dense SLAM Without A Pose Graph}.
\newblock In {\em Proc.~of Robotics: Science and Systems (RSS)}, 2015.

\bibitem{wu2024icra}
Y.~Wu, T.~Guadagnino, L.~Wiesmann, L.~Klingbeil, C.~Stachniss, and H.~Kuhlmann.
\newblock {LIO-EKF: High Frequency LiDAR-Inertial Odometry using Extended
  Kalman Filters}.
\newblock In {\em Proc.~of the IEEE Intl.~Conf.~on Robotics \& Automation
  (ICRA)}, 2024.

\bibitem{wei2022tro}
W.~Xu, Y.~Cai, D.~He, J.~Lin, and F.~Zhang.
\newblock {FAST-LIO2: Fast Direct LiDAR-Inertial Odometry}.
\newblock {\em IEEE Trans.~on Robotics (TRO)}, 38(4):2053--2073, 2022.

\bibitem{zhang2014rss}
J.~Zhang and S.~Singh.
\newblock {LOAM: Lidar Odometry and Mapping in Real-time}.
\newblock In {\em Proc.~of Robotics: Science and Systems (RSS)}, 2014.

\end{thebibliography}
\flushend
\IfFileExists{./certificate/certificate.tex}{
  \subfile{./certificate/certificate.tex}
}{}
\end{document}